\documentclass{article}
\usepackage{spconf,amsmath,graphicx,amssymb}
\usepackage{bbm}
\usepackage{algorithm}
\usepackage{algorithmic}
\usepackage{url}
\usepackage{subcaption}
\usepackage{graphicx}
\usepackage{bbm}
\usepackage{tikz}
\usepackage{pgfplots}
\usepackage{xcolor}
\usepackage{ulem}

\usepgfplotslibrary{groupplots}
\pgfplotsset{compat=1.12}

\usetikzlibrary{
    pgfplots.colorbrewer,
}
\pgfplotsset{
    cycle list/.define={my marks}{
        every mark/.append style={solid,fill=\pgfkeysvalueof{/pgfplots/mark list fill}},mark=*\\
        every mark/.append style={solid,fill=\pgfkeysvalueof{/pgfplots/mark list fill}},mark=square*\\
        every mark/.append style={solid,fill=\pgfkeysvalueof{/pgfplots/mark list fill}},mark=triangle*\\
        every mark/.append style={solid,fill=\pgfkeysvalueof{/pgfplots/mark list fill}},mark=diamond*\\
    },
}

\title{Rethinking Unsupervised Neural Superpixel Segmentation}
\name{Moshe Eliasof$^*$, Nir Ben Zikri$^*$, Eran Treister \thanks{$^*$ Equal contribution.}}
\address{Department of Computer Science \\ Ben-Gurion University of the Negev, Beer-Sheva, Israel}
\begin{document}
%
\maketitle
\begin{abstract}

Recently, the concept of unsupervised learning for superpixel segmentation via CNNs has been studied. 
Essentially, such methods generate superpixels by convolutional neural network (CNN) employed on a single image, and such CNNs are trained without any labels or further information. Thus, such approach relies on the incorporation of priors, typically by designing an objective function that guides the solution towards a meaningful superpixel segmentation.
In this paper we propose three key elements to improve the efficacy of such networks: (i) the similarity of the \emph{soft} superpixelated image compared to the input image, (ii) the enhancement and consideration of object edges and boundaries and (iii) a modified architecture based on atrous convolution, which allow for a wider field of view, functioning as a multi-scale component in our network. 
By experimenting with the BSDS500 dataset, we find evidence to the significance of our proposal, both qualitatively and quantitatively.
\end{abstract}
\section{Introduction}

\label{sec:intro}
The concept of superpixels suggests to group pixels that are locally similar, typically based on color-values and other low-level features. Often times superpixels serve as an alternative representation of images, which may ease the computations and improve performance on various tasks, e.g., semantic segmentation and optical flow prediction ~\cite{ssn2018}. The seminal works of SLIC ~\cite{slic}, SEEDS ~\cite{seeds}, FH ~\cite{graph_sp} suggest to consider the position and color proximity to generate superpixels.
Recently, it was proposed in a series of works ~\cite{tu2018learning, ssn2018, yang2020superpixel,wang2021ainet} to leverage CNNs for superpixel generation in a supervised, task-driven manner. While appealing, still, in many cases the shortage of labelled data prevents the utilization of CNNs for superpixels. To that end, it was also proposed in ~\cite{suzuki2020superpixel, edgeAwareUnsupervised2021} to predict superpixels in an unsupervised manner, based on the interpretation of CNNs as a prior ~\cite{dip}, and by designing a loss function that guides the network to output superpixels that are coherent with respect to the input image.

\begin{figure}
    \centering
    \begin{minipage}{.25\linewidth}
            \begin{subfigure}[t]{.9\linewidth}
                \includegraphics[width=\textwidth]{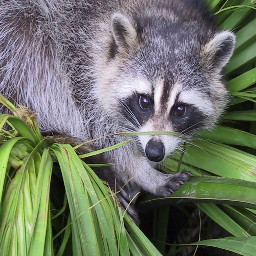}
                \caption{Input image}
                \label{fig:weather_activity}
            \end{subfigure}
        \end{minipage}
    \begin{minipage}{.25\linewidth}
        \begin{subfigure}[t]{.9\linewidth}
            \includegraphics[width=\textwidth]{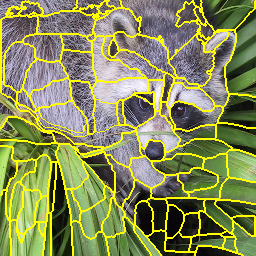}
            \label{fig:weather_filter1}
        \end{subfigure} \\
        \begin{subfigure}[b]{.9\linewidth}
            \includegraphics[width=\textwidth]{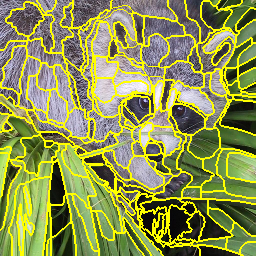}
            \caption{Superpixel segmentation}
            \label{fig:weather_filter2}
        \end{subfigure} 
    \end{minipage}
        \begin{minipage}{.25\linewidth}
        \begin{subfigure}[t]{.9\linewidth}
            \includegraphics[width=\textwidth]{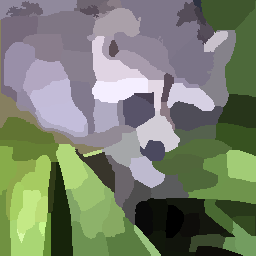}
            \label{fig:weather_filter1}
        \end{subfigure} \\
        \begin{subfigure}[b]{.9\linewidth}
            \includegraphics[width=\textwidth]{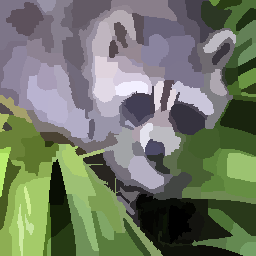}
            \caption{Superpixel image}
            \label{fig:weather_filter2}
        \end{subfigure} 
    \end{minipage}
    \caption{Illustration of the contribution of the soft superpixelated image $\hat{I}^p$ (bottom row) compared to reconstruction loss only with respect to $\tilde{I}$ (top row).}
        \label{fig:softVSnoSoft}
\end{figure}

In this paper we propose simple yet effective additions to the approach of ~\cite{suzuki2020superpixel, edgeAwareUnsupervised2021}, aiming to obtain superior quantitative metrics like achievable segmentation accuracy (ASA) and boundary recall (BR), as well as qualitatively better superpixel representation of images.
First, we propose to consider the image induced by the superpixel assignment learned by our network, which we refer to as the \textit{soft superpixelated image}. While the methods mentioned above demand the network to output additional 3 channels, aiming to reconstruct the input image by measuring the mean square error (MSE), we suggest to also compare the MSE with respect to the soft superpixelated image. We show that this addition is key to produce better superpixels in Fig. \ref{fig:softVSnoSoft}. In addition, since meaningful superpixels should be able to capture different objects in the image, we consider the edges and boundaries in the image similarly to ~\cite{edgeAwareUnsupervised2021}. To this end, we incorporate two important priors in our method. The first is the enhancement of edges features, obtained by extending our feature maps with their Laplacian response. This is motivated in Fig. \ref{fig:laplacian}, where it is evident that by considering the Laplacian of the feature maps, edges becomes more visible and thus guide our network to gravitate towards preserving edge features that exist in the input image, as depicted from Fig. \ref{fig:contourLossComparison}. Second, we introduce an objective function that considers the connection between edges found in the input image, its reconstruction from the network, and the differentiable soft superpixelated image.
Lastly, to improve the standard $3\times3$ convolutions in previous works  ~\cite{suzuki2020superpixel, edgeAwareUnsupervised2021}, we propose to employ atrous (dilated) convolutions in the form of an ASPP module from the DeepLabV3 architecture ~\cite{chen2017rethinking} in order to obtain a larger field of view, often used to solve the intimately connected image semantic segmentation task.

We verify the efficacy of our method by comparing with other baseline methods like ~\cite{slic,seeds,graph_sp,etps} as well as the recent \cite{suzuki2020superpixel, edgeAwareUnsupervised2021} using the BSDS500 ~\cite{bsd} dataset.

\section{Method}
\subsection{Notations}
Throughout this manuscript we denote an RGB input image by $I\in\mathbb{R}^{H\times W\times 3}$, where $H$ and $W$ denote the image height and width, respectively.
We denote by $N$ the maximal number of superpixels, which is a hyperparameter of our method. 

Our goal is to assign every pixel in the image $I$ to a superpixel. Thus, we may treat the pixel assignment prediction as a classification problem. To this end, we denote by
$P\in\mathbb{R}^{H\times W\times N}$ a probabilistic representation of the superpixels. The superpixel to which the $(i, j)$-th pixel belongs is given by the hard assignment of $s^{i,j} = \mathrm{arg}\max_sP_{i,j,s}$. Thus, we  define value of the $(i, j)$-th pixel of the \textit{hard superpixelated image} $I^{P}$ as follows:
\begin{equation}
    \label{eq:hardSuperPixImage}
    I^{P}_{i, j} = \frac{\sum_{h,w} \mathbbm{1}_{s^{i,j} = s^{h,w}}I_{h,w}}{\sum_{h,w} \mathbbm{1}_{s^{i,j} = s^{h,w}}},
\end{equation}
where $\mathbbm{1}_{s^{i,j} = s^{h,w}}$ is an indicator function that reads 1 if the hard assignment of the $(i, j)$-th and $(h, w)$-th pixel is equal.

\subsection{Soft assignment of superpixels}
\label{sec:SoftAssignment}
An important characteristic of a superpixel algorithm is its ability to obtain a superpixelated image that adheres to the input image $I$. This is often ~\cite{slic} obtained by considering the similarity (e.g., by MSE means) between the two. However, this procedure is not differentiable (due to the hard assignment of superpixels), a trait that is required to train CNNs. 
Therefore, we propose to consider the differentiable \textit{soft superpixelated image}, which at the $(i, j)$-th pixel reads
\begin{equation}
    \label{eq:softSuperPixImage}
    \hat{I}^{P}_{i, j} = \frac{\sum_{s=1}^{N} \sum_{h,w} P_{h,w,s}I_{h,w}}{\sum_{s=1}^{N}\sum_{h,w} P_{h,w,s}}.
\end{equation}
In the following, we utilize $\hat{I}^{P}$ in our objective functions.

\begin{figure}
    \centering
    \begin{minipage}{.25\linewidth}
        \begin{subfigure}[t]{.9\linewidth}
            \includegraphics[width=\textwidth]{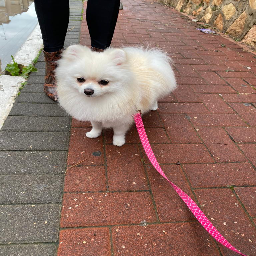}
            \label{fig:weather_filter1}
        \end{subfigure} \\
        \begin{subfigure}[b]{.9\linewidth}
            \includegraphics[width=\textwidth]{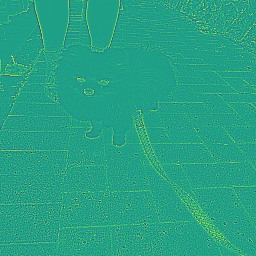}
            \caption{$I$}
            \label{fig:weather_filter2}
        \end{subfigure} 
    \end{minipage}
    \begin{minipage}{.25\linewidth}
        \begin{subfigure}[t]{.9\linewidth}
            \includegraphics[width=\textwidth]{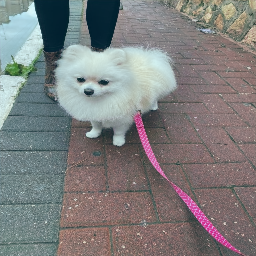}
            \label{fig:weather_filter1}
        \end{subfigure} \\
        \begin{subfigure}[b]{.9\linewidth}
            \includegraphics[width=\textwidth]{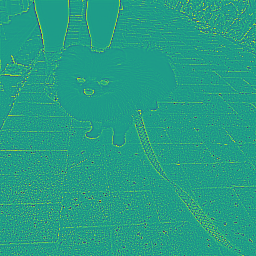}
            \caption{$\tilde{I}$}
            \label{fig:weather_filter2}
        \end{subfigure} 
    \end{minipage}
        \begin{minipage}{.25\linewidth}
        \begin{subfigure}[t]{.9\linewidth}
            \includegraphics[width=\textwidth]{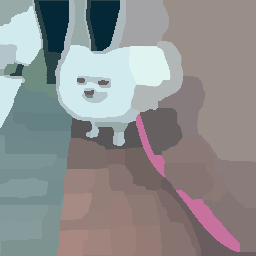}
            \label{fig:weather_filter1}
        \end{subfigure} \\
        \begin{subfigure}[b]{.9\linewidth}
            \includegraphics[width=\textwidth]{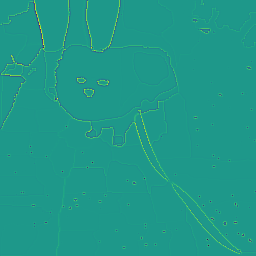}
            \caption{$\hat{I}^P$}
            \label{fig:weather_filter2}
        \end{subfigure} 
    \end{minipage}
    \caption{The application of the Laplacian kernel (bottom row) to a the input image, the reconstructed image $\tilde{I}$ and the soft superpixelated image $\hat{I}^P$ (top row).}
        \label{fig:laplacian}
\end{figure}

\begin{figure}
    \centering
    \begin{minipage}{.3\linewidth}
        \begin{subfigure}[t]{1\linewidth}
            \includegraphics[width=\textwidth]{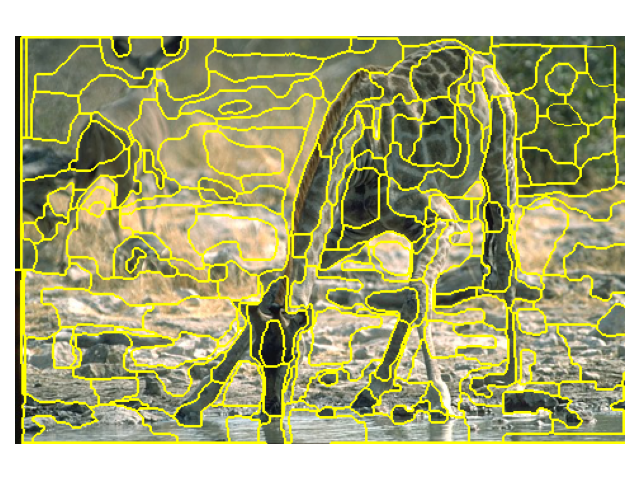}
            \label{fig:weather_filter1}
            \caption{Without edge awareness}

        \end{subfigure}
    \end{minipage}
    \begin{minipage}{.3\linewidth}
        \begin{subfigure}[t]{1\linewidth}
            \includegraphics[width=\textwidth]{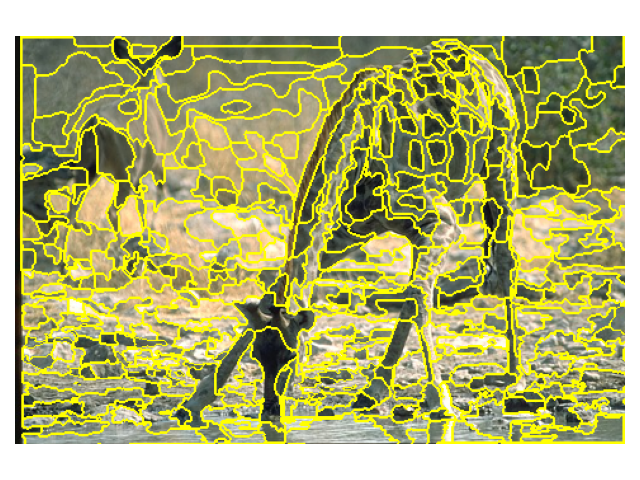}
            \label{fig:weather_filter1}
            \caption{With edge awareness}
        \end{subfigure}
    \end{minipage}
    \caption{The effect of edge awareness by adding Laplacian feature maps and $\mathcal{L}_{edge}$.}
        \label{fig:contourLossComparison}
\end{figure}

\subsection{Edge awareness}
\label{sec:edgeAwareness}
Superpixel algorithms can be compared by their ability to meaningfully segment an image, typically separating between different objects. Some of the key elements that define objects in images are their boundaries and edges.
Here, we propose an addition to the observations previously made in \cite{edgeAwareUnsupervised2021}, suggesting to consider edge and boundaries in the input image $I$.
Namely, we show that our network can benefit from considering the edges of the learned feature maps, and by constraining the relationship between the edges formed in our soft superpixelated image $\hat{I}^P$ and the input image $I$.
In order to find edges in the considered feature maps, we propose to utilize the Laplacian operator. An example of the proposed edge images to consider are presented in Fig. \ref{fig:laplacian}. Our approach differs from \cite{edgeAwareUnsupervised2021} in two major points. First, we propose a different objective function as explained in Sec. \ref{sec:Objectives}. Second, we propose to utilize Laplacian feature maps in our network in Sec. \ref{sec:architecture}.

\subsection{Objective functions}
\label{sec:Objectives}
Our objective function builds on the concept of regularized information maximization (RIM) for image superpixel generation presented in ~\cite{suzuki2020superpixel}, and the edge awareness idea presented in ~\cite{edgeAwareUnsupervised2021}.
The objective function is comprised of the following:
\begin{align}
    \nonumber
    \label{eq:allLosses}
    \mathcal{L}_\mathrm{objective} =& \mathcal{L}_\mathrm{clustering}+\alpha\mathcal{L}_\mathrm{smoothness}\\ &+ \beta\mathcal{L}_{recon} + \eta\mathcal{L}_{edge} ,
\end{align}
where $\mathcal{L}_\mathrm{clustering}$,   $\mathcal{L}_\mathrm{smoothness}$, and $\mathcal{L}_{recon}$ denote the RIM loss presented in ~\cite{suzuki2020superpixel} (and based on ~\cite{rim,mut_info}), spatial smoothness, and the reconstruction loss, respectively.
We balance the different terms using $\alpha$, $\beta$ and $\eta$, which are hyper-parameters. We now present each of the discussed objectives.

The objective $\mathcal{L}_\mathrm{clustering}$ is defined as follows:
\begin{align}
\nonumber
    \mathcal{L}_\mathrm{clustering}=&\frac{1}{HW}\sum_{i,j}\sum_{s=1}^{N}-P_{i,j,s}\log P_{i,j,s}\\
    &+\lambda\sum_{s=1}^{N}\hat{P}_{s}\log\hat{P}_{s},
\end{align}
where $\hat{P}_s=\frac{1}{HW}\sum_{i,j}P_{i,j,s}$ denotes the mean probability of the $s$-th superpixels, over all pixels.
The first term describes the per-pixel entropy $P_{i,j}\in\mathbb{R}^N$, pushing the predicted assignment vector to be deterministic.
The second term is the negative entropy of the mean vector over all pixels, which promotes the area-uniformity of the superpixels.

Next, the objective $\mathcal{L}_{smoothness}$ measures the difference between neighboring pixels, by utilizing the popular ~\cite{monodepth, suzuki2020superpixel} smoothness prior that seeks correspondence between smooth regions in the image and the predicted assignment tensor $P$:
\begin{align}
    \nonumber
    \label{eq:basicSmooth}
    \mathcal{L}_\mathrm{smoothness}=\frac{1}{HW}\sum_{i,j}\left(\left\|\partial_xP_{i,j}\right\|_1e^{-\|\partial_x I_{i,j}\|_2^2/\sigma}\right.\\
    +\left.\left\|\partial_yP_{i,j}\right\|_1e^{-\|\partial_y I_{j,j}\|_2^2/\sigma}\right),
\end{align}
where $P_{i,j}\in\mathbb{R}^N$ and $I_{i,j}\in\mathbb{R}^3$ are the features of the $(i, j)$-th pixel of $P$ and $I$, respectively. $\sigma$ is a scalar set to 8. 

We also add a reconstruction term to our objective.
First, we follow the standard reconstruction objective by adding 3 output channels denoted by $\tilde{I} \in \mathbb{R}^{H \times W \times 3}$ to the network, and minimizing
\begin{align}
    \frac{1}{3HW}\sum_{i,j}\|I_{i,j}-\tilde{I}_{i,j}\|_2^2.
\end{align}
While the justification for using this objective function, as discussed in ~\cite{suzuki2020superpixel} is to promote the network towards learning detail preserving convolutional filters, we claim that in order to obtain meaningful superpixels, it is also important to require similarity between the predicted  $\hat{I}^{P}$ and the input image $I$. Therefore our reconstruction loss is given by
\begin{align}
    \label{eq:recons}
    \mathcal{L}_\text{recons}=\frac{1}{3HW} (\sum_{i,j}\|I_{i,j}-\tilde{I}_{i,j}\|_2^2 +  \sum_{i,j}\|I_{i,j}-\hat{I}^{P}_{i,j}\|_2^2).
\end{align}

Fig. \ref{fig:softVSnoSoft} demonstrates the impact of the addition of the second term in $\mathcal{L}_\text{recons}$, where it is evident that a more coherent superpixelated image is obtained.

Finally, as discussed in Sec. \ref{sec:edgeAwareness}, we also consider the edges found in the input image $I$, the soft superpixelated image $\hat{I}^P$ and the reconstructed image $\tilde{I}$ and employ the Kullback–Leibler (KL) divergence loss to match between their edge distributions. We obtain the edges maps by computing the response of the images with a $3\times3$ Laplacian kernel followed by a Softmax to obtain values in $[0, 1]$, and denote them $E_I$, $E_{\tilde{I}}$ and $E_{\hat{I}^P}$, respectively.
We define our edge loss by
\begin{align}
    \label{eq:edge}
    \mathcal{L}_\text{edge}= KL(E_{I}, E_{\tilde{I}}) + KL(E_{I}, E_{\hat{I}^P}).
\end{align}

\begin{figure*}[t]
    \centering
    \begin{tabular}{ccccccc}
        \begin{minipage}{0.125\hsize}
        \centering
        \includegraphics[width=1\hsize]{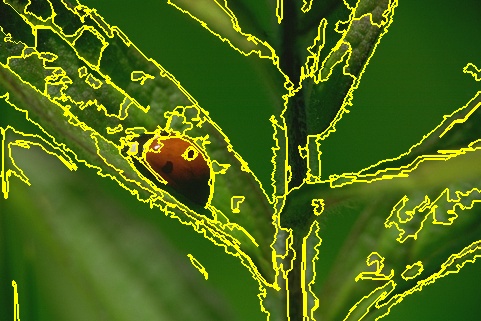}
        \end{minipage}
        \begin{minipage}{0.125\hsize}
        \centering
        \includegraphics[width=1\hsize]{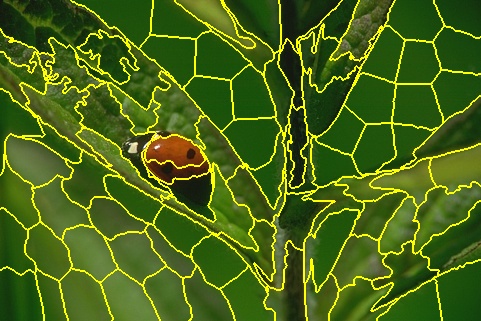}
        \end{minipage}
        \begin{minipage}{0.125\hsize}
        \centering
        \includegraphics[width=1\hsize]{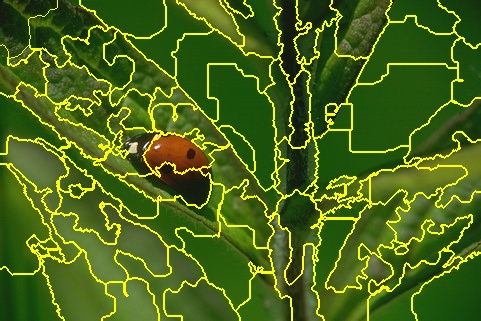}
        \end{minipage}
        \begin{minipage}{0.125\hsize}
        \centering
        \includegraphics[width=1\hsize]{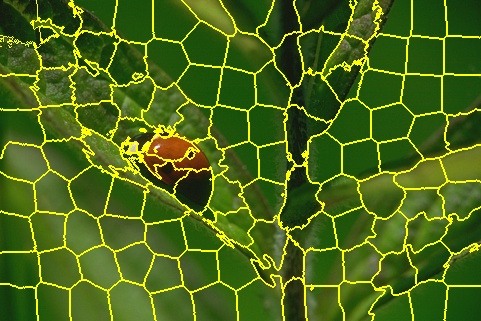}
        \end{minipage}
        \begin{minipage}{0.125\hsize}
        \centering
        \includegraphics[width=1\hsize]{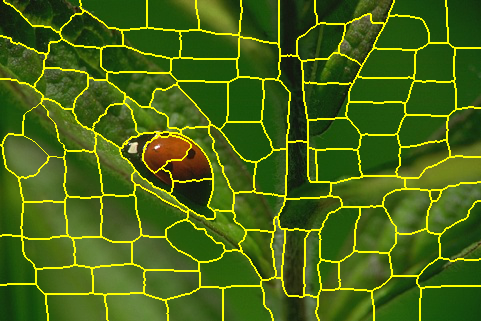}
        \end{minipage}
        \begin{minipage}{0.125\hsize}
        \centering
        \includegraphics[width=1\hsize]{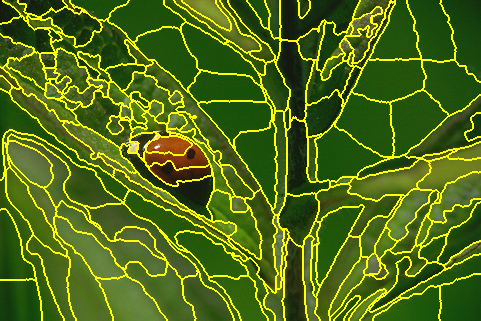}
        \end{minipage}
        \begin{minipage}{0.125\hsize}
        \centering
        \includegraphics[width=1\hsize]{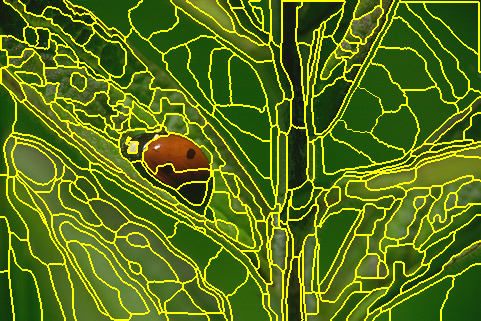}
        \end{minipage}
        \\ \\
        \begin{minipage}{0.125\hsize}
        \centering
        \includegraphics[width=1\hsize]{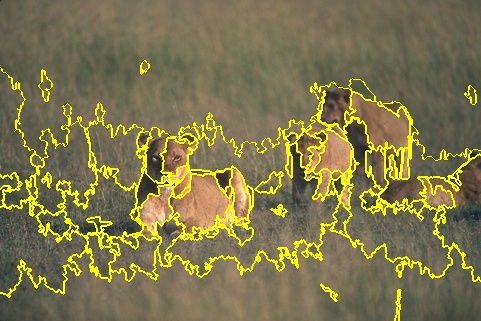}
        (a)
        \end{minipage}
        \begin{minipage}{0.125\hsize}
        \centering
        \includegraphics[width=1\hsize]{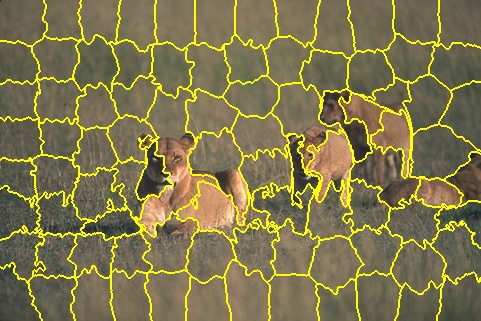}
        (b)
        \end{minipage}
        \begin{minipage}{0.125\hsize}
        \centering
        \includegraphics[width=1\hsize]{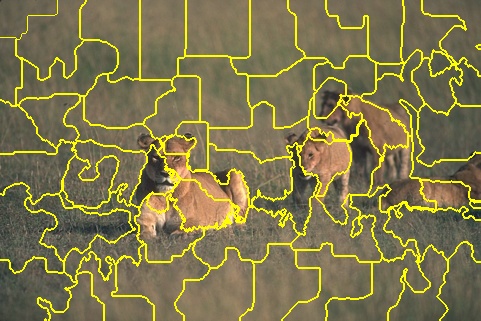}
        (c)
        \end{minipage}
        \begin{minipage}{0.125\hsize}
        \centering
        \includegraphics[width=1\hsize]{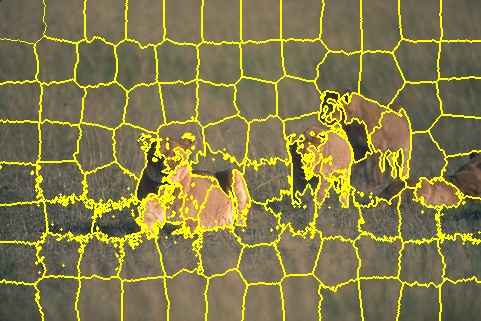}
        (d)
        \end{minipage}
        \begin{minipage}{0.125\hsize}
        \centering
        \includegraphics[width=1\hsize]{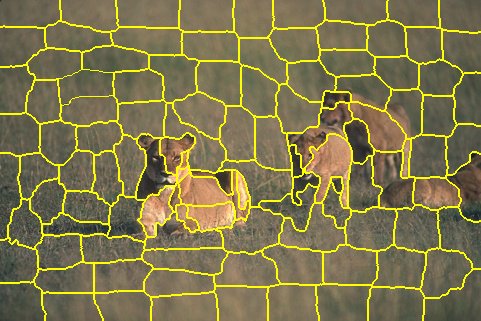}
        (e)
        \end{minipage}
        \begin{minipage}{0.125\hsize}
        \centering
        \includegraphics[width=1\hsize]{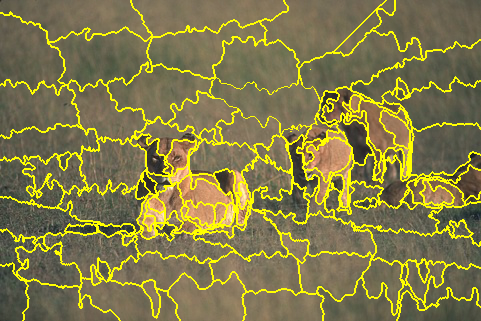}
        (f)
        \end{minipage}
        \begin{minipage}{0.125\hsize}
        \centering
        \includegraphics[width=1\hsize]{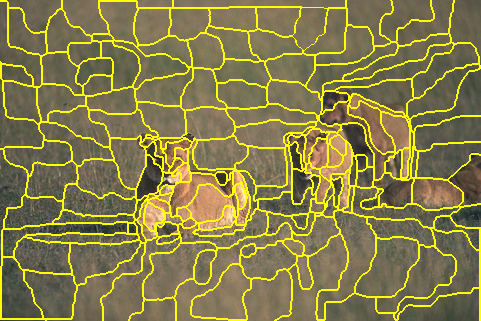}
        (i)
        \end{minipage}
    \end{tabular}
    \caption{Example results of (a) FH (b) SLIC (c) SEEDS (d) LSC (e) RIM (f) EA and (i) ours with 100 superpixels.}
    \label{fig:results_images}
\end{figure*}

\begin{figure*}
\centering
\begin{subfigure}[t]{.48\linewidth}
    \centering
    \begin{tikzpicture}
      \begin{axis}[
          width=1.0\linewidth, 
          height=0.60\linewidth,
          grid=major,
          grid style={dashed,gray!30},
          xlabel=Superpixels,
          ylabel=ASA score,
          ylabel near ticks,
          legend style={at={(0.0,0.0)},anchor=north,scale=0.8, draw=none, cells={anchor=west}, font=\tiny, fill=none},
          legend columns=6,
          xtick={25,50,100,200,400},
          xticklabels = {25,50,100,200,400},
          yticklabel style={
            /pgf/number format/fixed,
            /pgf/number format/precision=3
          },
          scaled y ticks=false,
          ymin = 0.77, ymax = 0.97,
          every axis plot post/.style={thick},
        ]
        \addplot[red, mark=*]
        table[x=superpixels,y=etps,col sep=comma] {figs_data/asa.csv};
        \addplot[blue, mark=*]
        table[x=superpixels,y=seeds,col sep=comma] {figs_data/asa.csv};
        \addplot[yellow,mark=*]
        table[x=superpixels,y=slic  ,col sep=comma] {figs_data/asa.csv};
        \addplot[magenta,mark=*]
        table[x=superpixels,y=fh,col sep=comma] {figs_data/asa.csv};
        \addplot[purple,mark=*]
        table[x=superpixels,y=lsc,col sep=comma] {figs_data/asa.csv};
        \addplot[green,mark=*]
        table[x=superpixels,y=rim,col sep=comma] {figs_data/asa.csv};
        \addplot[cyan,mark=*]
        table[x=superpixels,y=edge,col sep=comma] {figs_data/asa.csv};
        \addplot[black,mark=*]
        table[x=superpixels,y=ours,col sep=comma] {figs_data/asa.csv};

        \end{axis}
    \end{tikzpicture}
    \end{subfigure}
    \begin{subfigure}[t]{.48\linewidth}
    \centering
    \begin{tikzpicture}
      \begin{axis}[
          width=1.0\linewidth, 
          height=0.60\linewidth,
          grid=major,
          grid style={dashed,gray!30},
          xlabel=Superpixels,
          ylabel= BR score,
          ylabel near ticks,
          legend style={at={(0.7,0.01)},anchor=south,scale=1.0, draw=none, cells={anchor=east}, font=\normalsize, fill=none},
          legend columns=2,
          xtick={25,50,100,200,400},
          xticklabels = {25,50,100,200,400},
          yticklabel style={
            /pgf/number format/fixed,
            /pgf/number format/precision=3
          },
          scaled y ticks=false,
          every axis plot post/.style={thick},
        ]
        \addplot[red, mark=*]
        table[x=superpixel,y=etps,col sep=comma] {figs_data/br.csv};
        \addplot[blue, mark=*]
        table[x=superpixel,y=seeds,col sep=comma] {figs_data/br.csv};
        \addplot[yellow, mark=*]
        table[x=superpixel,y=slic,col sep=comma] {figs_data/br.csv};
        \addplot[magenta, mark=*]
        table[x=superpixel,y=fh,col sep=comma]
        {figs_data/br.csv};
        \addplot[purple,mark=*]
        table[x=superpixel,y=lsc,col sep=comma] {figs_data/br.csv};
        \addplot[green,mark=*]
        table[x=superpixel,y=rim,col sep=comma] {figs_data/br.csv};
        \addplot[cyan,mark=*]
        table[x=superpixel,y=edge,col sep=comma] {figs_data/br.csv};
        \addplot[black, mark=*]
        table[x=superpixel,y=ours,col sep=comma] {figs_data/br.csv};
    
        \legend{ETPS, SEEDS, SLIC, FH, LSC, RIM, EA, Ours}
        \end{axis}
    \end{tikzpicture}
    
    \end{subfigure}%
\caption{A comparison of our methods with others via the Achievable segmentation accuracy (ASA) and boundary recall (BR) metrics on the BSDS500 dataset as a function of the number of superpixels.}
\label{fig:bsds500}
\end{figure*}
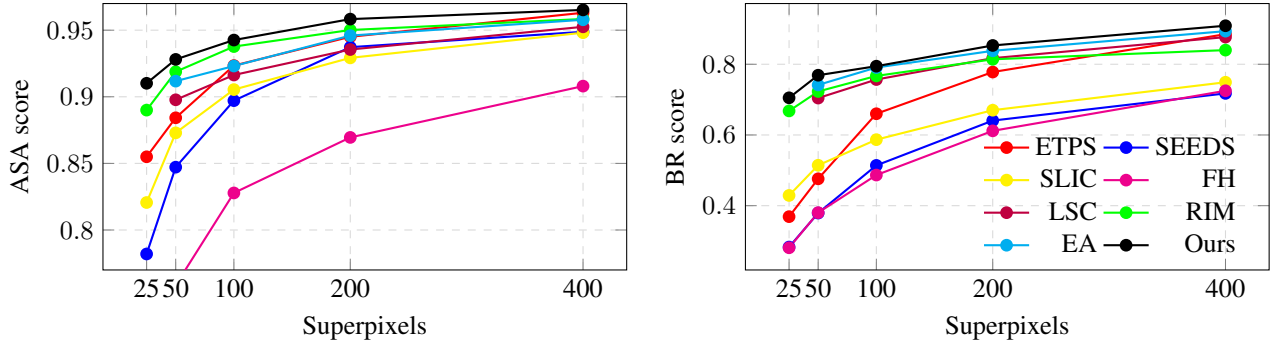

\subsection{Network architecture}
\label{sec:architecture}

\begin{figure}[H]
    \centering
    \includegraphics[width=0.4\textwidth]{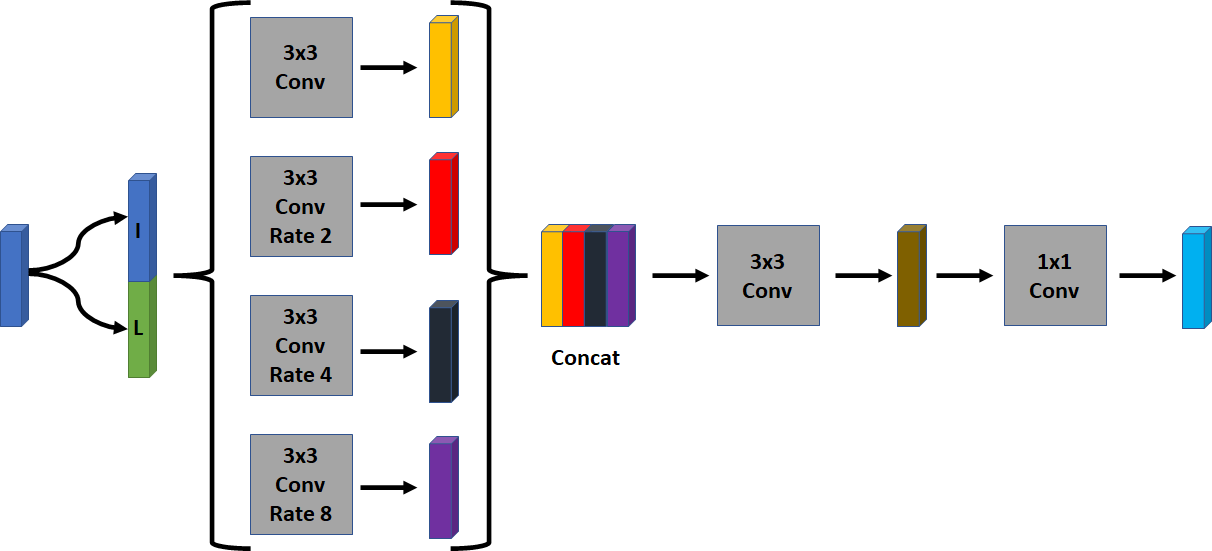}
    \caption{Our ASPP module and projection layers. The input of the module are the feature maps from the first 4 ConvInReLU blocks. I and L denote identity and Laplacian, respectively.}
    \label{fig:arch}
\end{figure}

Our network is comprised of two main components. The first is treated as a feature extractor, which is a sequence of convolutional layers with kernel of $3\times3$, instance normalization and the ReLU activation functions (dubbed as ConvInReLU), and the number of channels is doubled past every layer, starting from 32 channels. In our experiments, we use 4 ConvInReLU layers.
The output of the latter is first concatenated with its response to a $3\times3$ Laplacian kernel, and fed into the network described in Fig. \ref{fig:arch}, beginning with an ASPP ~\cite{chen2017rethinking} module with 4 atrous convolutions with variable dilation rates, in order to obtain a multi-scale combination of features, for an enhanced  field of view. In our implementation we use dilation rates of 1,2,4 and 8 to find compact and connected superpixels.
Finally, we perform a \emph{projection} step consisting of a single ConvInReLU layer, followed by a $1\times1$ convolution, in order to transform the output features to the desired shape of $[P, \tilde{I}] \in \mathbb{R}^{H \times W \times (N+3)}$. 
We also apply the Softmax function to $P$ to obtain valid probability distributions of the superpixel assignment for each pixel.

\section{Experiments}
We start by qualitatively inspecting the different components of our network. We present in Fig. \ref{fig:softVSnoSoft} the superpixel image obtained when adding $\hat{I}^P$ to the reconstruction objective $\mathcal{L}_{recons}$. It is apparent that the obtained superpixelated image is more similar to the input image when incorporating $\hat{I}^{P}$ in the objective function. In Fig. \ref{fig:laplacian} we present an example of the considered Laplacian maps, and their effect on the predicted superpixels in Fig. \ref{fig:contourLossComparison}. Then, we compare our method with various algorithms like SLIC~\cite{slic}, SEEDS~\cite{seeds}, ETPS ~\cite{etps}, FH~\cite{graph_sp} and LSC \cite{lsc2015}, as well as the RIM ~\cite{suzuki2020superpixel} and Edge aware (EA) ~\cite{edgeAwareUnsupervised2021} on the Berkeley Segmentation Dataset and Benchmarks (BSDS500)~\cite{bsd}. Since our method is unsupervised, we only utilize the 200 test images for the evaluation in the experiments.
Our network is implemented in PyTorch~\cite{pytorch}. For comparison with other methods, we use the publicly available resources from OpenCV\cite{opencv} and scikit-image ~\cite{skimage} and the author's ~\cite{suzuki2020superpixel, edgeAwareUnsupervised2021} code.

\subsection{Metrics}
We use the achievable segmentation accuracy (ASA), which describes the segmentation upper bound. Also, we employ the boundary recall (BR), with boundary tolerance of $r=2$. The BR metric compares the predicted boundaries with those in the ground truth segmentation images in our data. \footnote{Detailed discussion and definitions can be found in \cite{suzuki2020superpixel, ssn2018}.} 

\subsection{Implementation details}
The overall architecture details are discussed in Sec. \ref{sec:architecture}. 
We optimize the model for 1,000 iterations using Adam~\cite{adam}, with a learning rate of $0.01$, without weight-decay.
The balancing coefficients $(\lambda, \alpha, \beta)$ are set to $(2,2,10, 1)$ in our experiments, and they were selected by a naive grid search.

We follow the approach of ~\cite{suzuki2020superpixel} and as an additional input, we feed the network with the pixels locations $X\in\mathbb{R}^{H\times W\times 2}$ as an input. Thus, the input of the network is a tensor of shape $H\times W\times 5$\, which we also normalize such that each channel has a mean and variance of 0 and 1.

\subsection{Results}
Fig.~\ref{fig:bsds500} shows that our method achieves similar or better results compared to other methods, from classic to recent deep learning based methods,both in ASA and BR metrics.
We show qualitative examples of various methods Fig.~\ref{fig:results_images}. It is observed that our model is capable of capturing more details compared to other methods, and tends to produce superpixels that are in congruence with the boundaries of the objects in the image.

\section{conclusion}
We propose a soft superpixelated image as a medium to optimize our objective function to produce meaningful superpixels, together with the reinforcement of edge awareness, both in our objective function and the features of our network. We also suggest to employ a modified ASPP module to achieve multiscale properties, for an enhanced field of view.
We qualitatively reason about the various components of our method in Figs. \ref{fig:softVSnoSoft}--\ref{fig:contourLossComparison} and show their positive influence on the predicted superpixelated image.
Quantitatively, we experiment with BSDS500 and find that our method yields superior results compared to other methods in terms of ASA and BR.

\bibliographystyle{IEEEbib}
\bibliography{strings,refs}

\end{document}